\pdfoutput=1

\documentclass[11pt]{article}

\usepackage{emnlp2021}
\usepackage{times}
\usepackage{epsfig}
\usepackage{graphicx}
\usepackage{amsmath}
\usepackage{amssymb}
\usepackage{booktabs}
\usepackage{multirow}
\usepackage{multicol}
\usepackage{bm}
\usepackage{colortbl}
\definecolor{na}{gray}{0.9}

\usepackage{color,soul}
\usepackage{microtype}      
\usepackage{subcaption}
\usepackage{algorithm}
\usepackage{algpseudocode}%

\usepackage{times}
\usepackage{latexsym}

\usepackage[T1]{fontenc}

\usepackage{microtype}
\usepackage{bm}
\usepackage{soul}
\usepackage{url}
\usepackage{graphicx}
\usepackage{amsmath}
\usepackage{booktabs}
\usepackage{microtype}
\usepackage{graphicx}
\usepackage{booktabs} 
\usepackage{amsmath}
\usepackage{booktabs}
\usepackage{graphics}
\usepackage{graphicx}
\usepackage{epstopdf}
\usepackage{caption}
 \usepackage{wrapfig}
\usepackage{enumitem}
\usepackage{psfrag}
\usepackage{amssymb}
\usepackage{amsbsy}
\usepackage{graphicx}
\usepackage{array}
\usepackage{multirow}
\usepackage{bm}
\usepackage{verbatim}
\usepackage{relsize}
\usepackage{algorithm}
\usepackage{algpseudocode}
\usepackage{tabularx}
 \usepackage{verbatim}
\usepackage{bm}
\usepackage{lipsum}
\usepackage{tabularx}
\usepackage{mwe} 
\usepackage{url}            
\usepackage{booktabs}       
\usepackage{nicefrac}       
\usepackage{microtype}      
\usepackage{times}
\usepackage{epsfig}
\usepackage{graphicx}
\usepackage{amsmath}
\usepackage{amsthm}
\usepackage{mathrsfs} 
\usepackage{amssymb}
\usepackage{wrapfig}
\usepackage{lscape}
\usepackage{rotating}

\usepackage{wrapfig}

\usepackage{bm}

\usepackage{setspace}
\usepackage{xspace}

\title{Domain Adaptation for Sentiment Analysis Using Increased Intraclass Separation}

 \author{  Mohammad Rostami   \quad Aram Galstyan  \\
   USC Information Sciences Institute
\\
 \texttt{\{mrostami,galstyan\}@isi.edu} 
}

\begin{document}
\maketitle
\begin{abstract}
Sentiment analysis is a costly yet necessary task for enterprises to study the opinions of their customers to improve their products   and to determine optimal   marketing strategies. Due to the existence of a wide range of domains across different products and services, cross-domain sentiment analysis methods have received significant attention. These methods mitigate the domain gap between different applications by training cross-domain generalizable classifiers which help to relax the need for   data annotation for each domain. Most existing methods focus on learning domain-agnostic representations that are invariant with respect to both the source and the target domains. As a result, a classifier that is trained using  the source domain annotated data   would generalize well in a related target domain. We introduce a new domain adaptation method which induces large margins between different classes in an embedding space. This embedding space is trained to be domain-agnostic by matching the data distributions across the domains. Large intraclass margins in the source domain help to reduce the effect of ``domain shift'' on the classifier performance  in the target domain. Theoretical
and empirical analysis are provided to demonstrate that the proposed method is effective.
\end{abstract}

\section{Introduction}
 The main goal in sentiment classification is to predict the polarity of users automatically after collecting
their feedback using online shopping platforms, e.g., Amazon customer reviews. Popularity of online shopping and reviews, fueled
further by the recent pandemic, provides a valuable resource for businesses
to study the behavior and preferences of consumers and to align their products and services with the
market demand. A major challenge for automatic sentiment analysis is that
polarity is expressed using completely dissimilar terms and phrases in different domains. For example, while terms
such as ``fascinating'' and ``boring'' are used to describe books,   terms such as ``tasty'' and ``stale'' are used to describe food products.  As a result of this   discrepancy, an NLP model that is trained for a particular domain may
not generalize well in other different domains, referred as the problem of ``domain gap''~\cite{wei2018person}. Since generating annotated training data for all domains is expensive and time-consuming~\cite{rostami2018crowdsourcing},
cross-domain sentiment analysis has gained significant attention recently~\cite{saito2017ASYMmetric,peng2018cross,barnes2018projecting,sarma2019shallow,li2019transferable,guo2020multi,du2020adversarial,gong2020unified,xi2020domain,dai2020adversarial,lin2020multi}. The goal in cross-domain sentiment classification is to relax the need for   annotation via transferring
knowledge from another domain in which annotated data is accessible to   train models for domains with unannotated data.

The above problem has been studied more broadly in the \textit{domain adaptation} (DA) literature~\cite{wang2018deep}. A common DA
approach    is to map data points  from two domains into a  shared embedding
space to align the distributions of both domains~\cite{redko2017theoretical,rostami2019learning}. Since the embedding space would become
domain-agnostic,  a source-trained classifier   will generalize well in
the target domain. In the sentiment analysis problem, this means that polarity of natural language can be
expressed independent of the domain in the embedding space to transcendent discrepancies. We can model this embedding space
using a shared deep encoder which is trained to align the distributions of both domains at its output space.
This training procedure have been implemented using either adversarial machine learning~\cite{pei2018multi,long2018conditional,li2019transferable,dai2020adversarial}, which aligns distributions indirectly, or by directly minimizing   loss functions that are designed to   align the   distributions of both domains~\cite{peng2018cross,barnes2018projecting,kang2019contrastive,rostami2019deep,guo2020multi,xi2020domain,lin2020multi,rostami2020sequential,stan2021unsupervised}.

\textbf{Contributions:} we develop a new cross-domain sentiment analysis algorithm
for model adaptation by increasing the margins between class-specific clusters in the source domain. Our idea is
based on learning a parametric distribution for the source domain   in a cross-domain embedding
space. We model this   distribution as a Gaussian
mixture modal (GMM). We estimate the GMM parameters  using a subset of source samples
for which the classifier is confident. As a result, larger margins between classes
are induced using the GMM distribution which help reducing domain gap. We then use this
parametric GMM distribution to align the source and the target distributions. We draw confident   samples from
this distribution and enforce the distribution of the target  domain matches the GMM distribution in addition
to the source   distribution in the embedding space. We provide theoretical analysis that our method
minimizes an upperbound for the target domain expected error. We also provide experiments and
  demonstrate that our algorithm outperforms   state-of-the-art   algorithms.

\section{Related Work}
While domain adaptation methods for visual domains     usually use generative adversarial networks (GANs)~\cite{goodfellow2014generative} and align distributions indirectly, the dominant approach for cross-domain sentiment analysis is to design appropriate loss functions that directly impose domain alignment. The main reason is that natural language is expressed in terms of discrete values such as words, phrases, and sentences. Since this domain is not continuous, even if we convert natural language into real-valued vectors, it will  not  be differentiable. Hence, adversarial learning procedure cannot be easily adopted for   natural language processing (NLP) applications. Several alignment loss functions have been designed for cross-domain sentiment analysis. A group of methods are based on aligning the low-order distributional moments, e.g., means and covariances,  in an embedding space~\cite{wu2016sentiment,peng2018cross,sarma2019shallow,guo2020multi}. An improvement over these methods is to use probability distribution metrics to consider the encoded information in the higher-order moments~\cite{courty2016optimal,shen2018wasserstein}. 
 Damodaran et al.~\cite{damodaran2018deepjdot} demonstrated that using Wasserstein distance (WD) for domain alignment  boosts the performance significantly in visual domains compared to aligning only the low-order distributional moments~\cite{long2015learning,sun2016deep}.
We rely on  the sliced Wasserstein distance (SWD) for aligning distribution. SWD has been used   for domain adaptation~\cite{lee2019sliced} and continual learning~\cite{rostami2019complementary,rostami2020generative} in visual domains due to having less computational load compared to WD.

The major reason for performance degradation of a source-trained model in a target domain stems from ``domain shift'', i.e., the boundaries between the classes change in the embedding space even for related domains which in turn increases possibility of misclassification. It has been argued that if an increased-margin classifier is trained in the source domain, it can generalize better than many methods that try to align distributions without further model adaptation~\cite{tommasi2013frustratingly}. Inspired by this argument, our method is based on both aligning distributions in the embedding space and also inducing larger margins between classes by learning a  ``parametric distribution'' for the source domain.  Our idea is based on the empirical observation that when a deep network  classifier is trained in a domain with annotated data, data points of classes form separable clusters in an embedding space, modeled via the network responses in hidden layers. This means that the source distribution can be modeled as a multimodal distribution in the embedding  space. We can estimate this multimodal distribution using a Gaussian mixture model (GMM). Our work is based on using the GMM distribution  to induce larger margins between the class-specific clusters after an initial training phase in the source domain. We estimate the GMM using the source sample for which the classifier is confident and use random samples with high-confident labels to induce larger margins between classes.

\section{Cross-Domain Sentiment Analysis}

 \begin{figure}[t!]
    \centering
    \includegraphics[width=\linewidth]{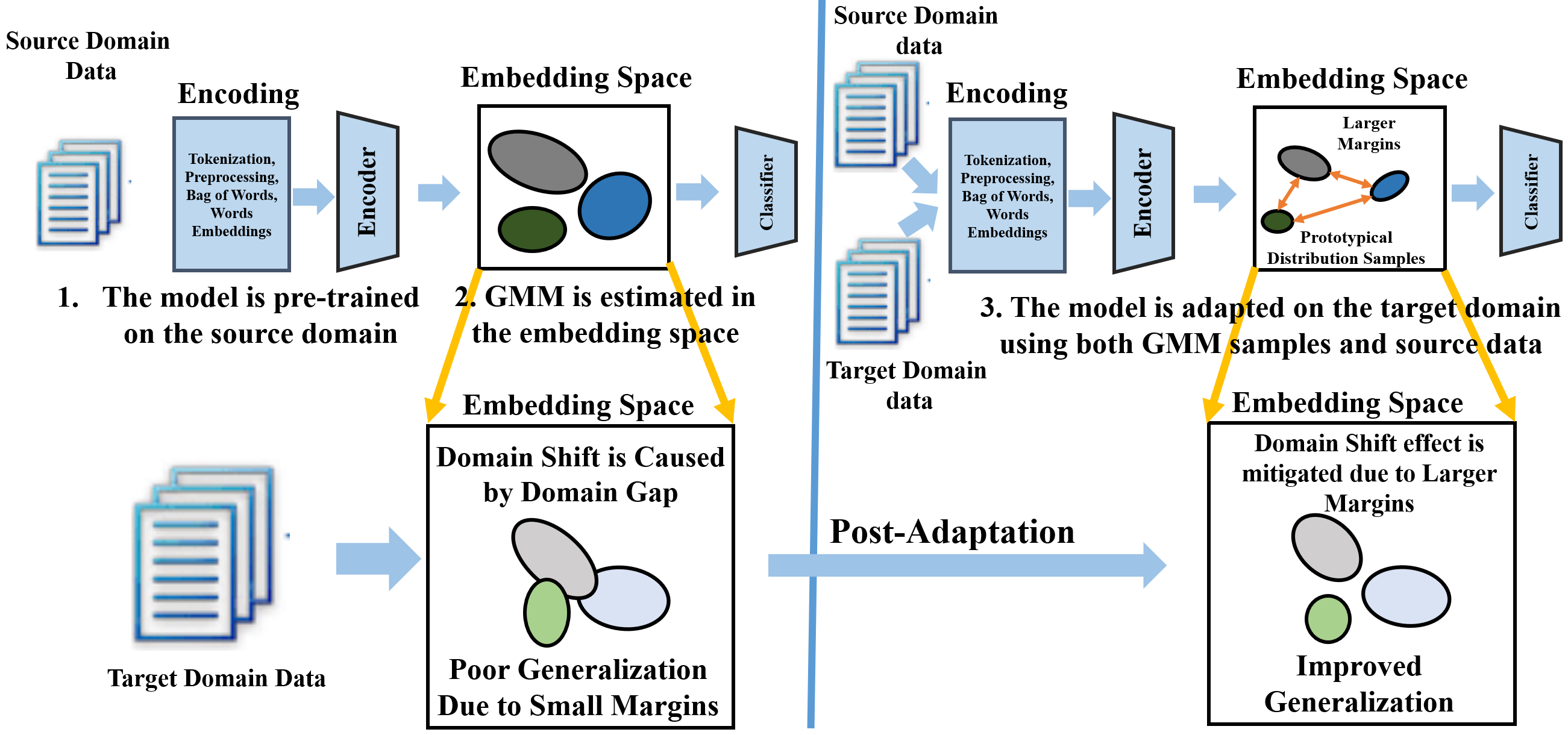}
         \caption{Architecture of the proposed cross-domain sentiment analysis framework. Left: separable clusters are formed in the embedding space after initial supervised model training   in the source domain and then   the multimodal distribution is estimated as a GMM. Right:    random samples from the   GMM with high-confident labels are used to generate a pseudo-dataset. The pseudo-dataset  helps to induce larger margins  between the classes to mitigate   domain shift in the target domain.  }
         \label{figDLSA:LL}
\end{figure}

Consider two sentiment analysis problems in a source domain $\mathcal{S}$ with an annotated dataset $D_\mathcal{S} = (\bm{X}_\mathcal{S},\bm{Y}_\mathcal{S})$, where $\bm{X}_{\mathcal{S}}=[\bm{x}_1^s,\ldots,\bm{x}_N^s]\in\mathcal{X}\subset\mathbb{R}^{d\times N}$   and   $\bm{Y}_{\mathcal{S}}=[\bm{y}^s_1,...,\bm{y}^s_N]\in \mathcal{Y}\subset\mathbb{R}^{k\times N}$, and a target domain $\mathcal{T}$ with an unannotated dataset $D_\mathcal{T} = (\bm{X}_\mathcal{S})$, where $\bm{X}_{\mathcal{T}}=[\bm{x}_1^t,\ldots,\bm{x}_N^t]\in\mathcal{X}\subset\mathbb{R}^{d\times M}$.  The real-valued feature vectors $\bm{X}_{\mathcal{S}}$ and $\bm{X}_{\mathcal{T}}$ are obtained after pre-processing the input text data using common  NLP methods, e.g., bag of words or word2vec. We consider that both domains share the same type of sentiments and hence the one-hot labels $\bm{y}_i^s$ encode $k$ sentiment types, e.g., negative or positive in binary sentiment analysis. Additionally, we    assume that  the source and the target feature data points are drawn independently and identically distributed    from the domain-specific distributions  $\bm{x}_i^s\sim p_{S}(\bm{x})$  and $\bm{x}_i^t\sim p_{T}(\bm{x})$. There is a gap between these distributions, i.e., $p_{T}(\bm{x})\neq p_{S}(\bm{x})$, which makes them distinct. 

Given  a   family of parametric functions $f_{\theta}:\mathbb{R}^d\rightarrow \mathcal{Y}$, e.g.,  deep neural networks with learnable weights $\theta$, and considering an ideal labeling function $f(\cdot)$, e.g., $\forall (\bm{x},\bm{y}): \bm{y} = f(\bm{x})$, the goal is to search for the optimal   predictor   $f_{\theta^*}(\cdot)$ in this family for the target domain. This model should have minimal expected error, i.e., $ \theta^* =\arg\min_{\theta}\{e_{\theta}\} =  \arg\min_{\theta}\{ \mathbb{E}_{\bm{x}^t\sim p_{T}(\bm{x})}(\mathcal{L}(f(\bm{x}^t),f_{\theta}(\bm{x}^t))\}$,  where $\mathcal{L}(\cdot)$ is a proper loss function and $\mathbb{E}(\cdot)$ denotes the expectation operator. Since the target domain data is unlabeled, the naive approach is to estimate the optimal model   using the standard empirical risk minimization (ERM) in the source domain:
\begin{equation}
\small
\begin{split}
 \hat{ \theta}&=\arg\min_{\theta}\{\hat{e}_{\theta}(\bm{X}_{\mathcal{S}},\bm{Y}_{\mathcal{S}},\mathcal{L})\}\\&=\arg\min_{\theta}\{\frac{1}{N}\sum_i \mathcal{L}(f_{\theta}(\bm{x}_i^s),\bm{y}_i^s)\}.
 \end{split}
 \label{eq:ERM}
\end{equation}
Given a large enough labeled data points in the source domain, ERM model  generalizes well in the source domain.   The source-trained model  may also perform much better than chance in the target domain, given cross-domain knowledge transfer. However, its performance will degrade in the target domain compared to its performance in the source domain because of existing distributional discrepancy between the two domains, since $p_{S}\neq p_{T}$.  Our goal is to benefit from the encoded information    in the unlabeled target domain data points and adapt the source-trained  classifier $f_{\hat{\theta}}$ to generalize   in the target domain. We use the common approach of reducing the domain gap across the domains by mapping data into a shared embedding space. 

We consider that the predictor model $f_\theta(\cdot)$  can be decomposed into  a deep encoder subnetwork $\phi_{\bm{v}}(\cdot): \mathcal{X}\rightarrow \mathcal{Z}\subset \mathbb{R}^p$ and a classifier subnetwork $h_{\bm{w}}(\cdot): \mathcal{Z}\rightarrow \mathcal{Y}$ such that $f_\theta = h_{\bm{w}}\circ \phi_{\bm{v}}$, where $\theta=(\bm{w},\bm{v})$. Here, $\mathcal{Z}$ is an embedding space which is modeled by the encoder responses at its output. We assume that the classes have become   separable for the source domain in this space after an initial training phase on the source domain (see Figure~\ref{figDLSA:LL}, left). If we can adapt the source-trained encoder network such that the two domains share similar distributions $\mathcal{Z}$, i.e., $\phi(p_{\mathcal{S}})(\cdot) \approx\phi(p_{\mathcal{T}})(\cdot)$, the embedding space would become domain-agnostic. As a result, the source-trained classifier network will generalize with similar performance in the target domain. 
 A number of prior cross-domain sentiment analysis algorithms use this strategy, select a proper probability distribution metric to compute the distance between the distributions $\phi(p_{\mathcal{S}}(\bm{x}^s))$ and $\phi(p_{\mathcal{T}}(\bm{x}^t))$, and then train the encoder network to align the domains via minimizing this distance:  
 \begin{equation}
 \small
\begin{split}
&\hat{\bm{v}},\hat{\bm{w}}=\arg\min_{\bm{v},\bm{w}} \frac{1}{N}\sum_{i=1}^N \mathcal{L}\big(h_{\bm{w}}(\phi_{\bm{v}}(\bm{x}_i^s)),\bm{y}_i^s\big)\\&+\lambda D\big(\phi_{\bm{v}}(p_{\mathcal{S}}(\bm{X}_{\mathcal{T}})),\phi_{\bm{v}}(p_{\mathcal{T}}(\bm{X}_{\mathcal{T}}))\big),
\end{split}
\label{eq:smallmainPrMatch}
\end{equation}  
 where $D(\cdot,\cdot)$ denotes a probability   metric to measure the domain discrepancy  and $\lambda$ is a trade-off parameter between the source ERM and the domain alignment term. We base our work on this general approach and use  SWD~\cite{lee2019sliced} to compute $D(\cdot,\cdot)$ in \eqref{eq:smallmainPrMatch}. Using SWD has three advantages. First, SWD can be computed efficiently compared to WD based on a closed form solution of WD distance in 2D. Second, SWD can be computed using the empirical samples that are drawn from the   two distributions. Finally, SWD possesses a non-vanishing gradient even when the support of the two distributions do not overlap~\cite{bonnotte2013unidimensional,lee2019sliced}.  Hence SWD is suitable for solving deep learning   problems which are normally handled using first-order gradient-based optimization techniques, e.g., Adam or SGD~\cite{stan2021privacy}.
 
 While   methods based on variations of Eq.~\eqref{eq:smallmainPrMatch} are effective to  reduce the domain gap to some extent, our goal is to improve upon the baseline obtained by Eq.~\eqref{eq:smallmainPrMatch} by introducing a loss term that increases the margins between classes in the source domain (check the embedding space in Figure~\ref{figDLSA:LL}, right, for having a better intuition). By doing so, our goal is to mitigate the negative effect of domain shift.

\section{Increasing Intraclass Margins}

 \begin{algorithm}[t]
\caption{$\mathrm{SAIM}^2\left (\lambda , \tau \right)$\label{ICLR2021Alg}} 
 { 
\begin{algorithmic}[1]
\State \textbf{Initial Training}: 
\State \hspace{2mm}\textbf{Input:} source dataset $\mathcal{D}_{\mathcal{S}}=(\bm{X}_{\mathcal{S}},  \bm{Y}_{\mathcal{S}})$,
\State \hspace{4mm}\textbf{Training on the Source Domain:}
\State \hspace{4mm} $\hat{ \theta}_0=(\hat{\bm{w}}_0,\hat{\bm{v}}_0) $
\State\hspace{9mm}$=\arg\min_{\theta}\sum_i \mathcal{L}(f_{\theta}(\bm{x}_i^s),\bm{y}_i^s)$
\State \hspace{2mm}  \textbf{GMM Distribution Estimation:}
\State \hspace{4mm} Use \eqref{eq:MAPestICLR} and estimate $\alpha_j, \bm{\mu}_j,$   $\Sigma_j$
\State \textbf{Model Adaptation}: 
\State \hspace{2mm} \textbf{Input:} target dataset $\mathcal{D}_{\mathcal{T}}=\bm{X}_{\mathcal{T}}$
\State \hspace{2mm} \textbf{Pseudo-Dataset Generation:} 
\State \hspace{4mm} $\mathcal{\hat{D}}_{\mathcal{P}}=(\textbf{Z}_{\mathcal{P}},\textbf{Y}_{\mathcal{P}})=$
\State \hspace{6mm} $([\bm{z}_1^p,\ldots,\bm{z}_N^p],[\bm{y}_1^p,\ldots,\bm{y}_N^p])$, where:
\State \hspace{16mm} $\bm{z}_i^p\sim \hat{p}_J(\bm{z}), 1\le i\le N_p$
\State \hspace{17mm}$\bm{y}_i^p=
\arg\max_j\{h_{\hat{\bm{w}}_0}(\bm{z}_i^p)\},$
\State \hspace{23mm}$\max\{h_{\hat{\bm{w}}_0}(\bm{z}_i^p)\}>\tau$
\For{$itr = 1,\ldots, ITR$ }
\State draw data batches from $\mathcal{ D}_{\mathcal{S}}$, $\mathcal{ D}_{\mathcal{T}}$, and $\mathcal{ D}_{\mathcal{P}}$
\State Update the model by solving \eqref{eq:mainPrMatchICLR}
\EndFor
\end{algorithmic}}
\label{algSAforUDA}
\end{algorithm} 

Our idea for increasing margins between the classes is based on learning an intermediate parametric distribution in the embedding space. We demonstrate that this distribution can be used to induce larger margins between the classes. To this end, we consider that the classifier subnetwork consists of a softmax layer. This means that the classifier should become  a maximum \textit{a posteriori} (MAP) estimator after training to be able to assign a membership probability to a given input feature vector. Under this formulation, the  model will generalize in the source domain if after supervised training of the model using the source data, the input distribution is transformed into a multi-modal distribution $p_J( \cdot)=\phi_{\bm{v}}(p_{\mathcal{S}})( \cdot)$  with $k$ modes in the embedding space (see Figure~\ref{figDLSA:LL}, left).
Each mode of this distribution corresponds to one type of sentiments.    
  The geometric distance between the  modes of this distribution corresponds to the margins between classes. If we test the source-trained model in the target domain, the boundaries between class modes will change due to the existence of ``domain shift'', i.e.,  $\phi_{\bm{v}}(p_{\mathcal{T}})( \cdot)\neq\phi_{\bm{v}}(p_{\mathcal{S}})( \cdot)$. Intuitively, as visualized in Figure~\ref{figDLSA:LL}, if we can increase the margins between the class-specific modes in the source domain, domain shift will likely cause less   performance degradation~\cite{tommasi2013frustratingly}.

We estimate the multimodal distribution in the embedding space  as a parametric GMM as follows:
\begin{equation}
\small
p_J(\bm{z})=\sum_{j=1}^k \alpha_j
\mathcal{N}(\bm{z}|\bm{\mu}_j,\bm{\Sigma}_j),
\end{equation}  
where  $\bm{\mu}_j$ and $\bm{\Sigma}_j$ denote the mean and co-variance matrices for each component and $\alpha_j$ denotes mixture weights for each component. We  solve for these parameters to estimate the multimodal distribution. 
Note that unlike usual cases in which iterative and time-consuming algorithms such as expectation maximization algorithm  need to be used for estimating the GMM parameters,  the source domain data points are labeled. As a result, we can estimate $\bm{\mu}_j$ and $\bm{\Sigma}_j$  for each component independently using standard MAP estimates. Similarly,  the   weights  $\alpha_j$ can be computed by a MAP estimate. Let $\bm{S}_j$ denote the support set for class $j$ in the training dataset, i.e., $\bm{S}_j=\{(\bm{x}_i^s,\bm{y}_i^s)\in \mathcal{D}_{\mathcal{S}}|\arg\max\bm{y}_i^s=j \}$. To cancel out outliers, we include only those source samples in the $\bm{S}_j$ sets, for which the source-trained model predicts the corresponding labels correctly. The closed-form MAP estimate for the mode parameters is given as:
\begin{equation}
\small
\begin{split}
&\hat{\alpha}_j = \frac{|\bm{S}_j|}{N},\hspace{2mm}\hat{\bm{\mu}}_j = \sum_{(\bm{x}_i^s,\bm{y}_i^s)\in \bm{S}_j}\frac{1}{|\bm{S}_j|}\phi_v(\bm{x}_i^s),\hspace{2mm}\\& \hat{\bm{\Sigma}}_j =\sum_{(\bm{x}_i^s,\bm{y}_i^s)\in \bm{S}_j}\frac{1}{|\bm{S}_j|}\big(\phi_v(\bm{x}_i^s)-\hat{\bm{\mu}}_j\big)^\top\big(\phi_v(\bm{x}_i^s)-\hat{\bm{\mu}}_j\big).
\end{split}
\label{eq:MAPestICLR}
\end{equation}    

Computations in Eq.~\eqref{eq:MAPestICLR} can be done efficiently. For a complexity analysis, please refer to the Appendix. Our   idea is to use this multimodal GMM distributional estimate to induce larger margins in the source domain (see Figure~\ref{figDLSA:LL}, right). We update the domain alignment term in \eqref{eq:smallmainPrMatch} to induce larger margins. To this end, we augment the source domain samples in the domain alignment term with samples of a labeled pseudo-dataset $\mathcal{D}_{\mathcal{P}}=(\textbf{Z}_{\mathcal{P}},\textbf{Y}_{\mathcal{P}})$ that we generate using the GMM estimate, where $\bm{Z}_{\mathcal{P}}=[\bm{z}_1^p,\ldots,\bm{z}_{N_p}^p]\in\mathbb{R}^{p\times N_p}, \bm{Y}_{\mathcal{P}}=[\bm{y}^p_1,...,\bm{y}^p_{N_p}]\in \mathbb{R}^{k\times N_p}$. This pseudo-dataset is  generated using the the GMM distribution. We   draw samples from the GMM distributional estimate $\bm{z}_i^p\sim \hat{p}_J(\bm{z})$ for this purpose. To induce larger margins between classes, we feed the initial drawn samples into the classifier network and check the confidence level of the classifier about its predictions for these randomly drawn  samples. We   set a confidence threshold $\tau\approx 1$ level and only  select a subset of the drawn samples for which the confidence level of the classifier is more than $\tau$:
\begin{equation}
\small
\begin{split}
&(\bm{z}_i^p,\bm{y}^p_i)\in \mathcal{D}_{\mathcal{P}}\hspace{2mm}  \text{if:}\hspace{2mm} \bm{z}_i^p\sim \hat{p}_J(\bm{z})\hspace{2mm}\text{and}\hspace{2mm}\\&  \max\{ h(\bm{z}_i^p)\}>\tau\hspace{2mm}\text{and}\hspace{2mm}  \bm{y}^p_i=\arg\max_i\{ h(\bm{z}_i^p)\}.
\end{split}
\label{eq:pesudosamples}
\end{equation}    
Given the GMM distributional form, selection of samples based on the threshold $\tau$ means that we include GMM samples that are close to the class-specific mode means $\mu_i$ (see Figure~\ref{figDLSA:LL}). In other words, the margins between the clusters in the source domain increase if we use the generated pseudo-dataset for domain alignment.   Hence, we update Eq.~\eqref{eq:smallmainPrMatch} and solve  the following problem:
\begin{equation}
\small
\begin{split}
&\hat{\bm{v}},\hat{\bm{w}}=\arg\min_{\bm{v},\bm{w}}\Big\{ \frac{1}{N}\sum_{i=1}^N \mathcal{L}\big(h_{\bm{w}}(\phi_{\bm{w}}(\bm{x}_i^s)),\bm{y}_i^s\big)\\&+\frac{1}{N_p}\sum_{i=1}^{N_p} \mathcal{L}\big(h_{\bm{w}}(\bm{z}_i^s),\bm{y}_i^s\big) +\lambda   \hat{D}\big(\phi_{\bm{v}}(\bm{X}_{\mathcal{T}}),\bm{X}_{\mathcal{P}}) )\Big)\\&+\lambda \hat{D}\big(\phi_{\bm{v}}(\bm{X}_{\mathcal{S}}), \bm{X}_{\mathcal{P}} \big)\big\} ,
\end{split}
\label{eq:mainPrMatchICLR}
\end{equation}  
 The first and the second terms in \eqref{eq:mainPrMatchICLR} are ERM terms for the source dataset and the generated pseudo-dataset in the embedding space to guarantee that the classifier continues to generalize well in the source domain after adaptation. The third and the fourth terms are empirical SWD losses   that align the source and the target domain distributions using   the pseudo-dataset which as we describe induces larger margins. The hope is that as visualized in Figure~\ref{figDLSA:LL}, these terms can reduce the effect of domain shift.  Our proposed solution, named   Sentiment Analysis using Increased-Margin Models    (SAIM$^2$), is presented  and visualized in Algorithm~\ref{ICLR2021Alg} and Figure~\ref{figDLSA:LL}.

\section{Theoretical Analysis}
 Following a standard PAC-learning framework~\cite{shalev2014understanding}, we prove that Algorithm~\ref{algSAforUDA} minimizes an upperbound for the target domain expected error.   Consider that the hypothesis class in a PAC-learning setting is the family of classifier sub-networks $\mathcal{H} = \{h_{\bm{w}}(\cdot)|h_{\bm{w}}(\cdot):\mathcal{Z}\rightarrow \mathbb{R}^k, \bm{v}\in \mathbb{R}^V\}$, where $V$ denotes the number of learnable parameters. We represent the expected error for a model $h_{\bm{w}}(\cdot)\in\mathcal{H} $ on the source and the target domains by $e_{\mathcal{S}}(\bm{w})$ and  $e_{\mathcal{T}}(\bm{w})$, respectively. Given the source and the target datasets, we can represent the empirical source and     target distributions in the embedding space as $\hat{\mu}_{\mathcal{S}}=\frac{1}{N}\sum_{n=1}^N\delta(\phi_{\bm{v}}(\bm{x}_n^s))$ and $\hat{\mu}_{\mathcal{T}}=\frac{1}{M}\sum_{m=1}^M\delta(\phi_{\bm{v}}(\bm{x}_m^t))$. Similarly, we can build an empirical distribution for the multimodal distribution $\hat{\mu}_{\mathcal{P}}=\frac{1}{N_p}\sum_{q=1}^{N_p}\delta(\bm{z}_n^q)$.  In our analysis we also use the notion of joint-optimal model $h_{\mathcal{S},\mathcal{T}}(\cdot)$ in our analysis which is defined as: $\bm{w}^*= \arg\min_{\bm{w}} e_{\mathcal{S},\mathcal{T}}=\arg\min_{\bm{w}}\{ e_{\mathcal{S}}+  e_{\mathcal{T}}\}$ for any given domains $\mathcal{S}$ and $\mathcal{T}$. When we have labeled data in both domains, this is  the best model that can be trained using ERM. Existence of a good joint-trained model guarantees that the domains are related, e.g., similar types of sentiment polarities are encoded   across the two  domains, to make positive knowledge transfer feasible.

\textbf{Theorem 1}: Consider that we use the procedure described in Algorithm~\ref{algSAforUDA} for cross-domain sentiment analysis, then the following inequality  holds for the target expected error:
\begin{equation}
\small
\begin{split}
e_{\mathcal{T}}\le & e_{\mathcal{S}} +\hat{D}(\hat{\mu}_{\mathcal{S}},\hat{\mu}_{\mathcal{P}})+\hat{D}(\hat{\mu}_{\mathcal{T}},\hat{\mu}_{\mathcal{P}})+(1-\tau)+e_{\mathcal{S},\mathcal{P}}\\&+\sqrt{\big(2\log(\frac{1}{\xi})/\zeta\big)}\big(\sqrt{\frac{1}{N}}+\sqrt{\frac{1}{M }}+2\sqrt{\frac{1}{N_p }}\big),
\end{split}
\label{eq:theroemforPLICLR}
\end{equation}    
where   $\xi$ is a constant which depends on  the properties of the loss function $\mathcal{L}(\cdot)$.

\textbf{Proof:} the complete proof  is included in  the Appendix.

Theorem~1 provides an explanation to justify Algorithm~\ref{algSAforUDA}. We observe that all the terms in the upperbound of the target expected error in   the right-hand side of \eqref{eq:theroemforPLICLR} are minimized by $SAIM^2$. The source expected error is minimized as the first term in \eqref{eq:mainPrMatchICLR}. The second and the third terms terms are minimized as the third and fourth terms of \eqref{eq:mainPrMatchICLR}. The fourth term $1-\tau$ will be small if we set $\tau\approx 1$. 
The term $e_{\mathcal{S},\mathcal{P}}$ is minimized through the first and the second term of \eqref{eq:mainPrMatchICLR}. This is highly important as using the pseudo-dataset provides a way to minimize this term. As can be seen in our proof in the Appendix, if we don't use the pseudo-dataset, this terms is replaced with $e_{\mathcal{S},\mathcal{T}}$ which cannot be minimized directly due to lack of having annotated data in the target domain.
   The last term in \eqref{eq:theroemforPLICLR} is a constant term that as common in PAC-learning can become negligible states that in order to train a good model if we have access to large enough datasets.  Hence all the terms in the upperbound are minimized and if this upperbound is tight,  then the process leads to training a good model for the target domain. If the two domain are related, e.g., share the same classes, and also classes become separable in the embedding space, i.e., GMM estimation error for the source domain distribution in the embedding space   is small, then the upperbound is going to be likely tight. However, we highlight that possibility of a tight upperbound is a condition for our algorithm to work. This is a common limitation for most parametric  learning algorithms.

\section{Experimental Validation}
  Our implemented code is available at \url{https://github.com/mrostami1366}.

\subsection{Experimental Setup}
We have selected the most common setup to perform our experiments for possibility of comparison.

\textbf{Dataset and preprocessing:} Most existing works report performance cross-domain tasks that are defined  using the Amazon Reviews benchmark dataset~\cite{blitzer2007biographies}. The  dataset is built using Amazon product reviews from four product domains:  Books (B), DVD (D), Electronics (E), and Kitchen (K)  appliances.  Each review is considered to have positive (higher than 3 stars) or negative (3 stars or lower) sentiment. Following the most common setup in the literature, we encode each review  in a 5000 dimensional or 30000 dimensional \textit{tf-idf} feature vector of bag-of-words unigrams and bigrams to compare our results against the existing benchmarks. Note that using more advanced text embedding methods   increases the absolute performance, but we are interested in studying the relative improvement compared to the source-trained model. For this reason, we use these features to have domain gap. Each   task  consists  of  2000  labeled     reviews for the source domain and 2000 unlabeled   reviews for the target domain, and   2500--5500 examples for testing. We report our performance on the 12 definable  cross-domain tasks for this dataset.  We report the average prediction accuracy and standard deviation (std) over 10 runs on the target domain testing split for our algorithm.

\textbf{Baselines}: We compare our method against   recently developed   algorithms. We compare against  DSN~\cite{bousmalis2016domain} CMD~\cite{zellinger2017central}, ASYM~\cite{saito2017ASYMmetric}, PBLM~\cite{ziser2018pivot},
MT-Tri~\cite{ruder2018strong}, TRL~\cite{ziser2019task}, and TAT~\cite{liu2019transferable}.  These works are representative of   advances in the field based on various approaches. 
DSN and CMD are similar to $SAIM^2$ in that both align distributions in an embedding space.
DSN learns  shared and domain-specific knowledge for each domain and aligns the shared knowledge using the mean-based maximum mean discrepancy metric. CMD uses the central moment discrepancy metric for domain alignment. ASYM benefits from the idea of pseudo-labeling of the target samples to updated the base model.
 MT-Tri is based on ASYM but it also benefits from multi-task learning. TRL and PBLM do not use distribution alignment and are based on the  pivot based language model. TAT is a recent work that has used adversarial learning successfully for cross-domain sentiment analysis. We   provided results by the authors for the tasks in our table.  We report std if std is reported in the original paper. All the methods except TAT that uses 30000 dimensional   features, use 5000 dimensional   features. Note that in the results, the methods are comparable if they use features with the same dimension for fair comparison.
  We    report performance of the source only (SO) model as a lowerbound to demonstrate the effect of adapting the model over this baseline. 

 \textbf{Model and optimization setup:} We used the benchmark   network architecture that is used in the above mentioned works for fair comparison. We used  an encoder with  one hidden dense layer  with 50   nodes with sigmoid activation function. The  classifiers  consist  of  a  softmax  layer  with two output nodes. We implemented our method in Keras, used adam optimizer   and tuned the learning rate in the source domain. We set $\tau = 0.99$  and $\lambda=10^{-2}$. We tuned them based on a brute-force search. We observed empirically that our algorithm is not sensitive   to the value of $\lambda$.   We   ran our code on a cluster   equipped with 4 Nvidia Tesla P100-SXM2 GPUs.

 \begin{table*}[t!]
 \centering 
{\small
\begin{tabular}{lc|cccccc}  
\hline
\hline
\multicolumn{2}{c}{Task}    & B$\rightarrow$D & B$\rightarrow$E & B$\rightarrow$K &D$\rightarrow$B &D$\rightarrow$E & D$\rightarrow$K   \\
\hline
\hline
\multicolumn{2}{c|}{SO}& 81.7  $\pm$ 0.2    & 74.0  $\pm$ 0.6   & 76.4   $\pm$ 1.0  &  74.5   $\pm$ 0.3    & 75.6    $\pm$ 0.7  &  79.5  $\pm$ 0.4    \\
\multicolumn{2}{c|}{DSN}& 82.8  $\pm$ 0.4   & 81.9  $\pm$ 0.5  &    84.4 $\pm$ 0.6  &  80.1   $\pm$ 1.3   &  81.4  $\pm$ 1.1  &  83.3  $\pm$    0.7 \\
\multicolumn{2}{c|}{CMD}& 82.6  $\pm$  0.3  & 81.5  $\pm$ 0.6  &    84.4 $\pm$ 0.3  &  $\bm{80.7}$   $\pm$ 0.6   &  82.2  $\pm$ 0.5   &    84.8 $\pm$ 0.2   \\
\multicolumn{2}{c|}{ASYM}&   80.7    &  79.8  &    82.5   &     73.2   &    77.0   &   82.5    \\
\multicolumn{2}{c|}{PBLM}&  84.2   &  77.6   & 82.5     &     82.5 &  79.6   &   83.2    \\
\multicolumn{2}{c|}{MT-Tri}&  81.2    &   78.0   &    78.8  &     77.1   &    81.0   &    79.5    \\
\multicolumn{2}{c|}{TRL}&   82.2 &   -   &    82.7   &     -    &    -   &    -   \\
\multicolumn{2}{c|}{$SAIM^2$}&  $\bm{83.2}$ $\pm$ 0.2   & $\bm{83.9}$  $\pm$ 0.3  &   $\bm{85.}9$ $\pm$ 0.3  &  80.3    $\pm$ 0.4   &  $\bm{84.2}$ $\pm$ 0.3  &    $\bm{87.3}$ $\pm$ 0.2   \\
\hline
\multicolumn{2}{c|}{TAT$^*$}&   84.5    &   80.1  &    83.6  &     81.9    &    81.9    &    84.0    \\
\multicolumn{2}{c|}{$SAIM^2$$^*$}&  86.2 $\pm$ 0.2   & 85.1  $\pm$ 0.2  &  87.6  $\pm$ 0.2  &  80.9   $\pm$ 0.5   & 85.2    $\pm$ 0.2   &  88.6  $\pm$ 0.2   \\
\hline
\hline
\multicolumn{2}{c}{Task}& E$\rightarrow$B & E$\rightarrow$D & E$\rightarrow$K &K$\rightarrow$B &K$\rightarrow$D & K$\rightarrow$E \\
\hline
\hline
\multicolumn{2}{c|}{SO}&  72.3 $\pm$    1.5 & 74.2  $\pm$ 0.6  &    85.6 $\pm$ 0.6  &  73.1   $\pm$ 0.1   &  75.2  $\pm$ 0.7  &  85.4  $\pm$    1.0 \\
\multicolumn{2}{c|}{DSN}& 75.1  $\pm$  0.4  &  77.1 $\pm$ 0.3  & 87.2    $\pm$ 0.7  &  76.4   $\pm$ 0.5   &  78.0  $\pm$ 1.4  & 86.7   $\pm$    0.7 \\
\multicolumn{2}{c|}{CMD}&  74.9 $\pm$ 0.6   &  77.4 $\pm$ 0.3  &    86.4 $\pm$ 0.9  &  75.8   $\pm$ 0.3   & 77.7   $\pm$ 0.4  & 86.7   $\pm$    0.6 \\
\multicolumn{2}{c|}{ASYM}&   73.2    &   72.9   &    86.9  &     72.5    &    74.9   &   84.6    \\
\multicolumn{2}{c|}{PBLM}& 71.4  & 75.0  & 87.8  & 74.2  & $\bm{79.8}$  & $\bm{87.1}$  \\
\multicolumn{2}{c|}{MT-Tri}&   73.5   &   75.4   &    87.2   &     73.8    &    77.8   &    86.0    \\
\multicolumn{2}{c|}{TRL}&   -   &   75.8   &    -   &    72.1    &    -   &    -   \\
\multicolumn{2}{c|}{$SAIM^2$}&  $\bm{78.6}$ $\pm$ 0.4   & $\bm{79.7}$  $\pm$  0.2  &  $\bm{89.2}$  $\pm$ 0.2  &   $\bm{76.7}$  $\pm$  0.4  &  79.1  $\pm$ 0.4  &    87.0 $\pm$  0.1  \\
\hline
\multicolumn{2}{c|}{TAT$^*$}&   83.2   &   77.9   &    90.0   &     75.8   &    77.7   &    88.2    \\
\multicolumn{2}{c|}{$SAIM^2$$^*$}&  78.8 $\pm$ 0.3   &  78.9 $\pm$ 0.3  &   90.1 $\pm$ 0.2  &   78.1  $\pm$  0.2  &    78.8 $\pm$  0.4 &  88.1  $\pm$ 0.1    \\
\hline
\hline
\end{tabular}}
\caption{ Classification accuracy for the cross-domain sentiment analysis tasks for Amazon Reviews dataset. In this table, $^*$ denotes methods that use $30000$ dimensional \textit{tf-idf} feature vectors.}
\label{table:tabDAICLR1}
\vspace{-4mm}
\end{table*}

\subsection{Comparative Results}

Our results   are reported in Table~\ref{table:tabDAICLR1}. In this table, bold font denotes best performance among the methods that use 5000 dimensional features. We see that $SAIM^2$ algorithm performs reasonably well and in most cases leads to the best performance. Note that this is not unexpected as none of the methods has the best performance across all tasks.    We observe from this table that overall the methods DSN and CMD which are based on aligning the source and target distributions- which are more similar to our algorithm-  have relatively similar performances. This observation suggests that we should not expect considerable performance boost if we simply align the distributions by  designing a new alignment loss function. This means that outperformance of $SAIM^2$ compared to these methods   stems from inducing larger margins. We verify this intuition in our ablative study.   We also observe that increasing the dimensional of \textit{tf-idf} features to 30000 leads to performance boost which is probably  the reason behind good performance of TAT. Hence, we need to use the same dimension for features for fair comparison among the methods.

\begin{figure}[tb!]
    \centering
           \begin{subfigure}[b]{0.215\textwidth}\includegraphics[width=\textwidth]{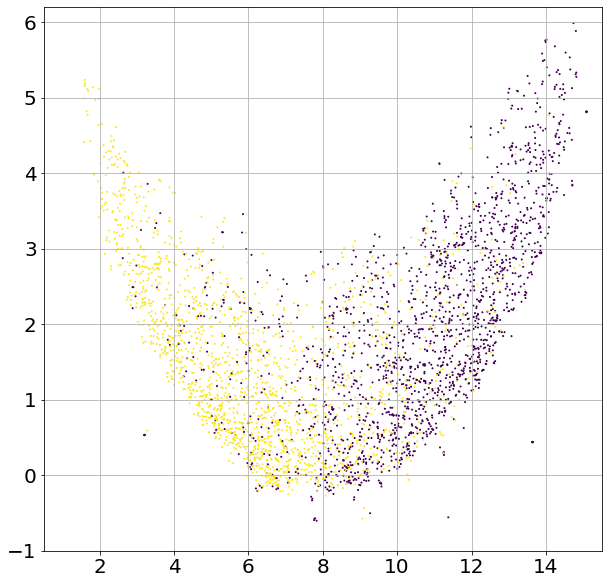}
           \centering
        \caption{Source}
        \label{NIPSDALfig:ICLRsource}
    \end{subfigure}
    \begin{subfigure}[b]{0.215\textwidth}\includegraphics[width=\textwidth]{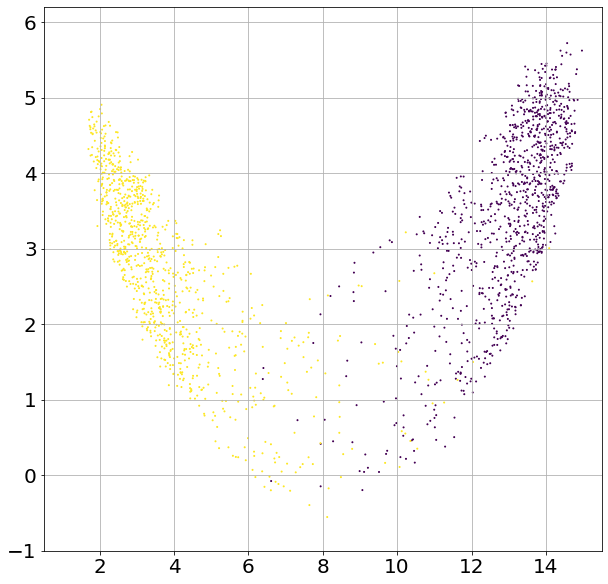}
           \centering
        \caption{GMM. Dist. Samples}
        \label{NIPSDALfig:ICLRGMM}
    \end{subfigure}
       \begin{subfigure}[b]{0.215\textwidth}\includegraphics[width=\textwidth]{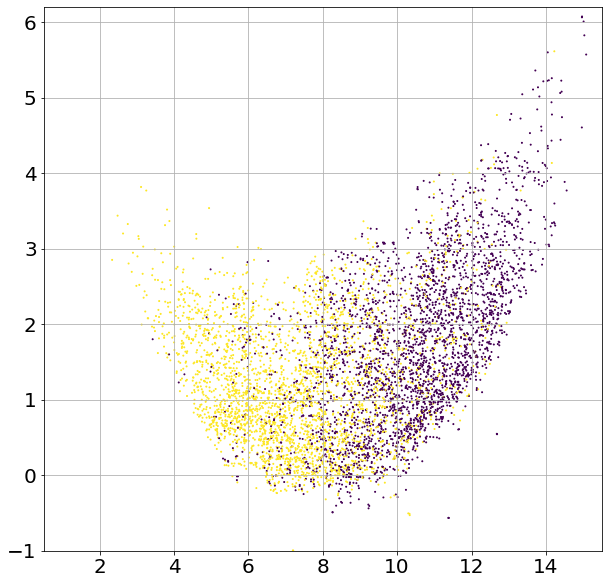}
           \centering
        \caption{Target: Pre-Adapt.}
        \label{NIPSDALfig:ICLRPre}
    \end{subfigure}
       \begin{subfigure}[b]{0.215\textwidth}\includegraphics[width=\textwidth]{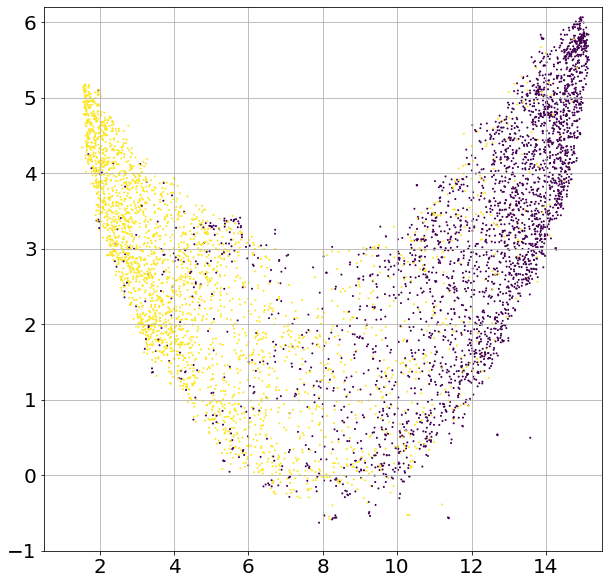}
              \centering
        \caption{Target: Post-Adapt.}
        \label{NIPSDALfig:ICLRpost}
    \end{subfigure}
     \caption{UMAP visualization for  the    D$\rightarrow$K task:  (a) the source domain   testing split, (b) the GMM distribution samples, (c) the target domain testing split pre-  and (d)   post-adaptation.  (Best viewed in color).  }\label{NIPSDALfig:ICLRembed}
     \vspace{-4mm}
\end{figure}

To provide an empirical exploration to validate the intuition  we used for our rationale, we  have used UMAP~\cite{mcinnes2018umap}   to reduce  the dimension of the data representations  in the 50D embedding space to two for the purpose of 2D visualization. Figure~\ref{NIPSDALfig:ICLRembed} visualizes the testing splits of the source domain before model adaptation, the testing splits of the target domain before and after model adaptation, and finally random samples drawn from the estimated GMM distribution  for the D$\rightarrow$K  task. Each point represents one data point and each color represents one of the sentiments. Observing Figure~\ref{NIPSDALfig:ICLRsource}  and Figure~\ref{NIPSDALfig:ICLRGMM}, we conclude that the estimated GMM   distribution approximates the source domain distribution reasonably well and at the same time, a margin between the classes in the boundary region  is observable. Figure~\ref{NIPSDALfig:ICLRPre} visualizes the target domain  samples prior to model adaptation. As expected, we observe that domain gap has caused less separations between the classes, as also evident from SO performance in Table~\ref{table:tabDAICLR1}. Figure~\ref{NIPSDALfig:ICLRpost} visualizes the target domain samples  after  adaptation using $SAIM^2$ algorithm. Comparing  Figure~\ref{NIPSDALfig:ICLRpost} with Figure~\ref{NIPSDALfig:ICLRPre} and Figure~\ref{NIPSDALfig:ICLRsource}, we see that the classes have become more separated. Also, careful comparison of Figure~\ref{NIPSDALfig:ICLRpost} and Figure~\ref{NIPSDALfig:ICLRGMM} reveals $SAIM^2$ algorithm has led to a bias in the target domain to move the data points further from the boundary.  These visualizations serve as an empirical verification for our theoretical analysis.

\subsection{Effect of Data Imbalance on Performance}
In practical setting,  the label distribution for the target domain training dataset cannot be enforced to be balanced due to absence of labels.   To study the effect of label imbalance using a controlled experiment, we synthetically design an imbalanced dataset using the Amazon dataset.  We designed two experiments, where the target domain datasets has the 90/10 and 80/20 ratios of imbalance between the two classes, respectively. We have provided domain adaptation results using for these two imbalanced scenarios in Table~\ref{table:tabDAICLR4}. We can see that performance of our algorithm  has degraded slightly but our algorithm has been robust 
to a large extent with respect to label imbalance.

\begin{table}[t!]
\setlength{\tabcolsep}{3pt}
 \centering 
{\tiny
\begin{tabular}{lc|cccccc}  
\hline
\hline
\multicolumn{2}{c}{Task}    & B$\rightarrow$D & B$\rightarrow$E & B$\rightarrow$K &D$\rightarrow$B &D$\rightarrow$E & D$\rightarrow$K   \\
\hline
\hline
\multicolumn{2}{c|}{$80/20$}&  82.8 $\pm$ 0.3   &   83.2 $\pm$ 0.5  &   85.5 $\pm$  0.3 &   78.7  $\pm$ 0.2    &  83.3  $\pm$ 0.2  &  86.8  $\pm$ 0.2    \\
\multicolumn{2}{c|}{$90/10$}& 82.9  $\pm$ 0.5   &  83.4 $\pm$ 0.3  &   85.8 $\pm$ 0.2  & 78.5    $\pm$ 0.4   &  83.3  $\pm$ 0.4  & 86.8   $\pm$ 0.3    \\
\hline  
\hline
\multicolumn{2}{c}{Task}& E$\rightarrow$B & E$\rightarrow$D & E$\rightarrow$K &K$\rightarrow$B &K$\rightarrow$D & K$\rightarrow$E \\
\hline
\hline
\multicolumn{2}{c|}{$80/20$}&  78.7 $\pm$ 0.2   & 78.5  $\pm$ 0.5  &   88.6 $\pm$ 0.1  &  76.3   $\pm$ 0.6    &  77.9  $\pm$ 0.4  & 86.6   $\pm$ 0.1    \\  
\multicolumn{2}{c|}{$90/10$}&  78.7 $\pm$ 0.2   & 78.0  $\pm$ 0.4  &  88.0  $\pm$ 0.2  &   76.5  $\pm$ 0.5   &  77.3  $\pm$ 0.3  &  86.7  $\pm$ 0.2    \\
\hline
\end{tabular}}
\caption{ Effect of label-imbalance on performance.}
\label{table:tabDAICLR4}
\end{table}

  \begin{figure}[tb!]
    \centering
           \begin{subfigure}[b]{0.215\textwidth}\includegraphics[width=\textwidth]{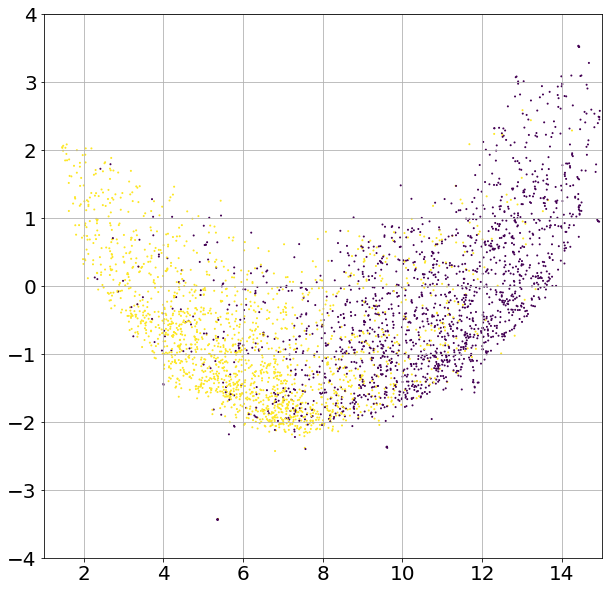}
           \centering
        \caption{Source}
        \label{NIPSDALfig:ICLRsource1}
    \end{subfigure}
    \begin{subfigure}[b]{0.215\textwidth}\includegraphics[width=\textwidth]{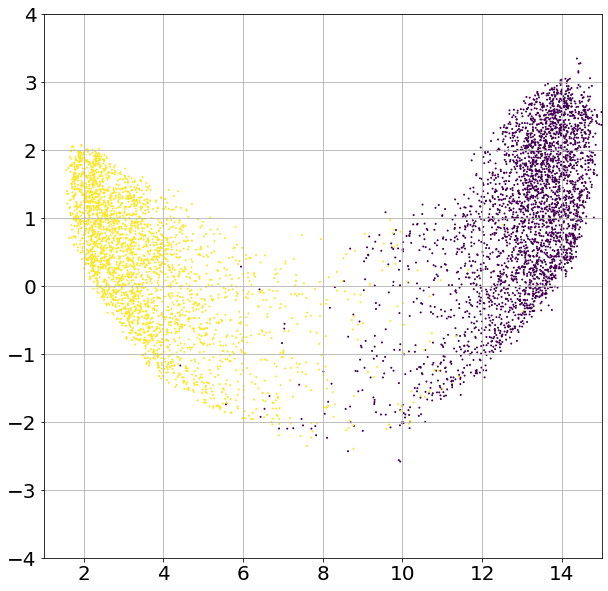}
           \centering
        \caption{GMM Dist. Samples}
        \label{NIPSDALfig:ICLRGMM1}
    \end{subfigure}
       \begin{subfigure}[b]{0.215\textwidth}\includegraphics[width=\textwidth]{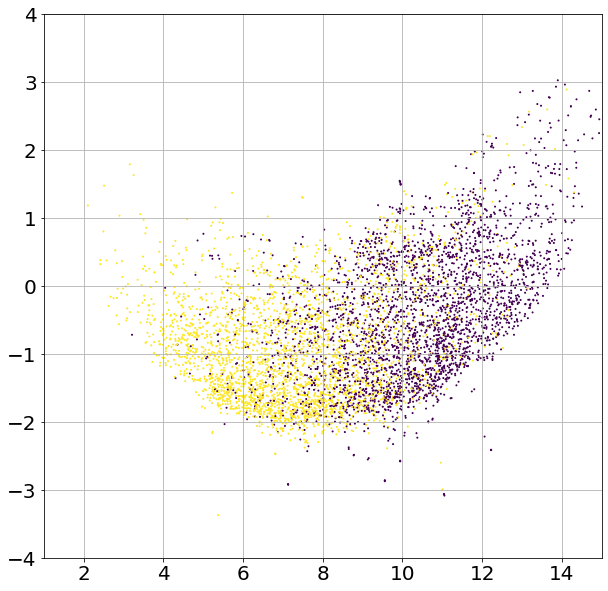}
           \centering
        \caption{Target: Pre-Adapt.}
        \label{NIPSDALfig:ICLRPre1}
    \end{subfigure}
       \begin{subfigure}[b]{0.215\textwidth}\includegraphics[width=\textwidth]{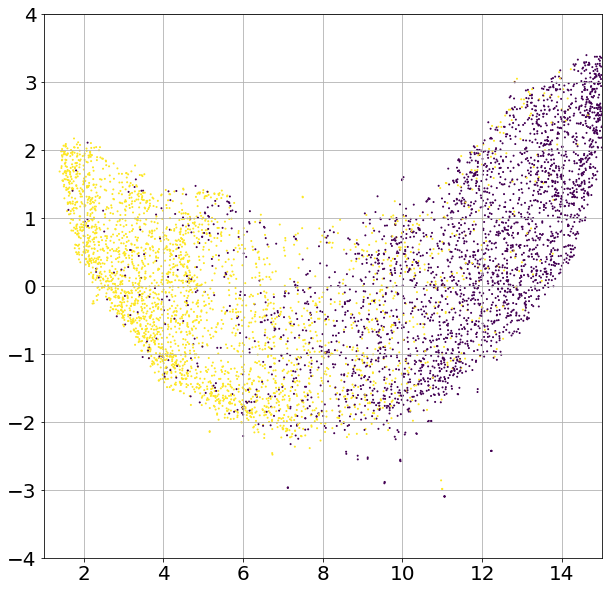}
              \centering
        \caption{Target: Post-Adapt.}
        \label{NIPSDALfig:ICLRpost1}
    \end{subfigure}
     \caption{UMAP visualization for  the  task D$\rightarrow$K task in the imbalanced regime of $90/10$:  (a) the source domain   testing split, (b) the GMM distribution samples, (c) the target domain testing split pre-, and (d) post-domain  adaptation.  (Best viewed in color).  }\label{NIPSDALfig:ICLRembed1}
     \vspace{-4mm}
\end{figure}

 For sanity check, we have presented the UMAP visualization for the testing split of the task D$\rightarrow$K for the   scenario     $90/10$  in Figure~\ref{NIPSDALfig:ICLRembed1}. Observations in Figure~\ref{NIPSDALfig:ICLRembed1} match what we reported in Table~\ref{table:tabDAICLR4}, confirming that our algorithm does require a balanced dataset imbalanced   is practically beneficial.

\subsection{Ablation Studies}
First note that the source only (SO) model result, which is trained using  \eqref{eq:ERM}, already serves as a basic ablative study to verify the effect of domain alignment. Improvement over this baseline demonstrates effect of domain adaptation on performance.

\begin{figure}[tb!]
\centering
\begin{minipage}[t]{0.4\linewidth}\vspace{0pt}
\includegraphics[width=.7\linewidth]{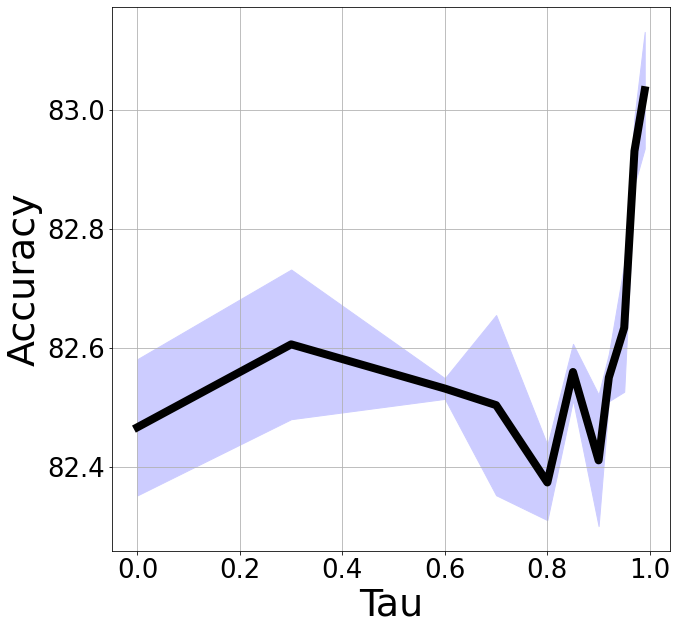}
\caption{Effect of $\tau$ on performance.}
\label{ICLR2021fig4}
\label{fig:model_selection}
\end{minipage}
\hfill
\begin{minipage}[t]{0.57\linewidth}
\captionof{table}{Performance for AO.}
\label{table:tabDAICLR2}
\resizebox{\linewidth}{!} 
{\footnotesize
\begin{tabular}{lc|ccc}  
\hline
\multicolumn{2}{c}{Task}    & B$\rightarrow$D & B$\rightarrow$E & B$\rightarrow$K     \\
\hline
\multicolumn{2}{c}{}&  81.9 $\pm$ 0.5    &  80.9 $\pm$ 0.8  &    83.2 $\pm$ 0.8     \\
\hline
\multicolumn{2}{c}{Task}      &D$\rightarrow$B &D$\rightarrow$E & D$\rightarrow$K   \\
\hline
\multicolumn{2}{c}{}&     74.0   $\pm$ 0.9    & 80.9    $\pm$ 0.7  & 83.4   $\pm$ 0.6    \\
\hline  
\multicolumn{2}{c}{Task}& E$\rightarrow$B & E$\rightarrow$D & E$\rightarrow$K   \\
\hline
\multicolumn{2}{c}{}&  74.1 $\pm$ 0.6    & 74.0  $\pm$ 0.3  &    87.8 $\pm$ 0.9     \\
\hline
\multicolumn{2}{c}{Task}    &D$\rightarrow$B &D$\rightarrow$E & D$\rightarrow$K   \\
\hline
\multicolumn{2}{c}{ }&  74.1 $\pm$ 0.6    & 74.0  $\pm$ 0.3  &    87.8 $\pm$ 0.9    \\
\hline
\end{tabular}}
\end{minipage}
 \vspace{-4mm}
\end{figure}

In Table~\ref{table:tabDAICLR2}, we have provided an additional ablative studies. 
We have reported result of alignment only (AO) model adaptation based on \eqref{eq:smallmainPrMatch}. The AO model does not benefit from the margins that $SAIM^2$ algorithm induces between the classes. Comparing AO results with Table~\ref{table:tabDAICLR1}, we can conclude that the effect of increased margins is important in our performance. Compared to other cross-domain sentiment analysis methods, the performance boost for our algorithm stems from inducing large margins. This suggests that researchers may check to investigate secondary techniques for domain adaptation in NLP domains, in addition to probability distribution alignment.

Finally, we have studied the effect of the value of the confidence parameter on performance.  In Figure~\ref{ICLR2021fig4}, we have visualized the performance of our algorithm for the task $B\rightarrow D$ when $\tau$ is varied in the interval $[0,0.99]$. When $\tau= 0$,   the samples are not necessarily   confident samples. We observe that as we increase the value of $\tau$, the performance increases as a result of inducing larger margins. For values $\tau>0.8$, the performance has less variance which suggests robustness of performance if $\tau\approx 1$.  These empirical observations about $\tau$ accord with our theoretical   result, stated in the upperbound   \eqref{eq:theroemforPLICLR}.

\section{Conclusions}
We  developed a method for cross-domain sentiment analysis  based on aligning two domain-specific distributions in a shared embedding space and inducing larger margins between the classes in the source domain using an intermediate multi-modal  GMM distribution. We   theoretically  justified our approach.  Our experiments demonstrate that our algorithm is effective.
  A  future research direction is to address cross-domain sentiment analysis when   different types of sentiments exists in two  domains. 
 
\bibliography{ref}

\begin{thebibliography}{47}
\expandafter\ifx\csname natexlab\endcsname\relax\def\natexlab#1{#1}\fi

\bibitem[{Barnes et~al.(2018)Barnes, Klinger, and
  im~Walde}]{barnes2018projecting}
Jeremy Barnes, Roman Klinger, and Sabine~Schulte im~Walde. 2018.
\newblock Projecting embeddings for domain adaption: Joint modeling of
  sentiment analysis in diverse domains.
\newblock In \emph{Proceedings of Comp. Linguistics}, pages 818--830.

\bibitem[{Bhushan~Damodaran et~al.(2018)Bhushan~Damodaran, Kellenberger,
  Flamary, Tuia, and Courty}]{damodaran2018deepjdot}
Bharath Bhushan~Damodaran, Benjamin Kellenberger, R{\'e}mi Flamary, Devis Tuia,
  and Nicolas Courty. 2018.
\newblock Deepjdot: Deep joint distribution optimal transport for unsupervised
  domain adaptation.
\newblock In \emph{Proceedings of the European Conference on Computer Vision
  (ECCV)}, pages 447--463.

\bibitem[{Blitzer et~al.(2007)Blitzer, Dredze, and
  Pereira}]{blitzer2007biographies}
John Blitzer, Mark Dredze, and Fernando Pereira. 2007.
\newblock Biographies, bollywood, boom-boxes and blenders: Domain adaptation
  for sentiment classification.
\newblock In \emph{Proceedings of ACL}, pages 440--447.

\bibitem[{Bolley et~al.(2007)Bolley, Guillin, and
  Villani}]{bolley2007quantitative}
Fran{\c{c}}ois Bolley, Arnaud Guillin, and C{\'e}dric Villani. 2007.
\newblock Quantitative concentration inequalities for empirical measures on
  non-compact spaces.
\newblock \emph{Probability Theory and Related Fields}, 137(3-4):541--593.

\bibitem[{Bonnotte(2013)}]{bonnotte2013unidimensional}
N.~Bonnotte. 2013.
\newblock \emph{Unidimensional and evolution methods for optimal
  transportation}.
\newblock Ph.D. thesis, Paris 11.

\bibitem[{Bousmalis et~al.(2016)Bousmalis, Trigeorgis, Silberman, Krishnan, and
  Erhan}]{bousmalis2016domain}
Konstantinos Bousmalis, George Trigeorgis, Nathan Silberman, Dilip Krishnan,
  and Dumitru Erhan. 2016.
\newblock Domain separation networks.
\newblock In \emph{Proceedings of NeurIPS}, pages 343--351.

\bibitem[{Courty et~al.(2016)Courty, Flamary, Tuia, and
  Rakotomamonjy}]{courty2016optimal}
Nicolas Courty, R{\'e}mi Flamary, Devis Tuia, and Alain Rakotomamonjy. 2016.
\newblock Optimal transport for domain adaptation.
\newblock \emph{IEEE transactions on pattern analysis and machine
  intelligence}, 39(9):1853--1865.

\bibitem[{Dai et~al.(2020)Dai, Liu, Ren, and Xu}]{dai2020adversarial}
Yong Dai, Jian Liu, Xiancong Ren, and Zenglin Xu. 2020.
\newblock Adversarial training based multi-source unsupervised domain
  adaptation for sentiment analysis.
\newblock In \emph{AAAI}.

\bibitem[{Du et~al.(2020)Du, Sun, Wang, Qi, and Liao}]{du2020adversarial}
Chunning Du, Haifeng Sun, Jingyu Wang, Qi~Qi, and Jianxin Liao. 2020.
\newblock Adversarial and domain-aware bert for cross-domain sentiment
  analysis.
\newblock In \emph{Proceedings of the 58th Annual Meeting of the Association
  for Computational Linguistics}, pages 4019--4028.

\bibitem[{Gong et~al.(2020)Gong, Yu, and Xia}]{gong2020unified}
Chenggong Gong, Jianfei Yu, and Rui Xia. 2020.
\newblock Unified feature and instance based domain adaptation for end-to-end
  aspect-based sentiment analysis.
\newblock In \emph{Proceedings of the 2020 Conference on Empirical Methods in
  Natural Language Processing (EMNLP)}, pages 7035--7045.

\bibitem[{Goodfellow et~al.(2014)Goodfellow, Pouget-Abadie, Mirza, Xu,
  Warde-Farley, Ozair, Courville, and Bengio}]{goodfellow2014generative}
Ian Goodfellow, Jean Pouget-Abadie, Mehdi Mirza, Bing Xu, David Warde-Farley,
  Sherjil Ozair, Aaron Courville, and Yoshua Bengio. 2014.
\newblock Generative adversarial nets.
\newblock In \emph{Advances in Neural Info. Proc. Systems}, pages 2672--2680.

\bibitem[{Guo et~al.(2020)Guo, Pasunuru, and Bansal}]{guo2020multi}
Han Guo, Ramakanth Pasunuru, and Mohit Bansal. 2020.
\newblock Multi-source domain adaptation for text classification via distance
  net-bandits.
\newblock In \emph{AAAI}, pages 7830--7838.

\bibitem[{Kang et~al.(2019)Kang, Jiang, Yang, and
  Hauptmann}]{kang2019contrastive}
Guoliang Kang, Lu~Jiang, Yi~Yang, and Alexander~G Hauptmann. 2019.
\newblock Contrastive adaptation network for unsupervised domain adaptation.
\newblock In \emph{Proceedings of CVPR}, pages 4893--4902.

\bibitem[{Lee et~al.(2019)Lee, Batra, Baig, and Ulbricht}]{lee2019sliced}
Chen-Yu Lee, Tanmay Batra, Mohammad~Haris Baig, and Daniel Ulbricht. 2019.
\newblock Sliced wasserstein discrepancy for unsupervised domain adaptation.
\newblock In \emph{Proceedings of the CVPR}, pages 10285--10295.

\bibitem[{Li et~al.(2019)Li, Li, Wei, Bing, Zhang, and
  Yang}]{li2019transferable}
Zheng Li, Xin Li, Ying Wei, Lidong Bing, Yu~Zhang, and Qiang Yang. 2019.
\newblock Transferable end-to-end aspect-based sentiment analysis with
  selective adversarial learning.
\newblock In \emph{Proceedings of EMNLP-IJCNLP}, pages 4582--4592.

\bibitem[{Lin et~al.(2020)Lin, Zhao, Meng, and Chua}]{lin2020multi}
Chuang Lin, Sicheng Zhao, Lei Meng, and Tat-Seng Chua. 2020.
\newblock Multi-source domain adaptation for visual sentiment classification.
\newblock In \emph{AAAI}, pages 2661--2668.

\bibitem[{Liu et~al.(2019)Liu, Long, Wang, and Jordan}]{liu2019transferable}
Hong Liu, Mingsheng Long, Jianmin Wang, and Michael Jordan. 2019.
\newblock Transferable adversarial training: A general approach to adapting
  deep classifiers.
\newblock In \emph{ICML}, pages 4013--4022.

\bibitem[{Long et~al.(2015)Long, Cao, Wang, and Jordan}]{long2015learning}
Mingsheng Long, Yue Cao, Jianmin Wang, and Michael Jordan. 2015.
\newblock Learning transferable features with deep adaptation networks.
\newblock In \emph{Proceedings of ICML}, pages 97--105.

\bibitem[{Long et~al.(2018)Long, Cao, Wang, and Jordan}]{long2018conditional}
Mingsheng Long, Zhangjie Cao, Jianmin Wang, and Michael~I Jordan. 2018.
\newblock Conditional adversarial domain adaptation.
\newblock In \emph{Proceedings of NeurIPS}, pages 1640--1650.

\bibitem[{McInnes et~al.(2018)McInnes, Healy, Saul, and
  Gro{\ss}berger}]{mcinnes2018umap}
Leland McInnes, John Healy, Nathaniel Saul, and Lukas Gro{\ss}berger. 2018.
\newblock {UMAP}: Uniform manifold approximation and projection.
\newblock \emph{Journal of Open Source Soft.}, 3(29):861.

\bibitem[{Moon(1996)}]{moon1996expectation}
Todd~K Moon. 1996.
\newblock The expectation-maximization algorithm.
\newblock \emph{IEEE Signal processing magazine}, 13(6):47--60.

\bibitem[{Pei et~al.(2018)Pei, Cao, Long, and Wang}]{pei2018multi}
Zhongyi Pei, Zhangjie Cao, Mingsheng Long, and Jianmin Wang. 2018.
\newblock Multi-adversarial domain adaptation.
\newblock In \emph{Proceedings Thirty-Second AAAI Conference}, pages
  3934--3941.

\bibitem[{Peng et~al.(2018)Peng, Zhang, Jiang, and Huang}]{peng2018cross}
Minlong Peng, Qi~Zhang, Yu-gang Jiang, and Xuan-Jing Huang. 2018.
\newblock Cross-domain sentiment classification with target domain specific
  information.
\newblock In \emph{Proceedings of ACL}, pages 2505--2513.

\bibitem[{Redko and Sebban(2017)}]{redko2017theoretical}
A.~Redko, I.and~Habrard and M.~Sebban. 2017.
\newblock Theoretical analysis of domain adaptation with optimal transport.
\newblock In \emph{Joint European Conference on Machine Learning and Knowledge
  Discovery in Databases}, pages 737--753. Springer.

\bibitem[{Rostami(2019)}]{rostami2019learning}
Mohammad Rostami. 2019.
\newblock \emph{Learning Transferable Knowledge Through Embedding Spaces}.
\newblock Ph.D. thesis, University of Pennsylvania.

\bibitem[{Rostami and Galstyan(2020)}]{rostami2020sequential}
Mohammad Rostami and Aram Galstyan. 2020.
\newblock Sequential unsupervised domain adaptation through prototypical
  distributions.
\newblock \emph{arXiv e-prints}, pages arXiv--2007.

\bibitem[{Rostami et~al.(2018)Rostami, Huber, and
  Lu}]{rostami2018crowdsourcing}
Mohammad Rostami, David Huber, and Tsai-Ching Lu. 2018.
\newblock A crowdsourcing triage algorithm for geopolitical event forecasting.
\newblock In \emph{Proceedings of the 12th ACM Conference on Recommender
  Systems}, pages 377--381. ACM.

\bibitem[{Rostami et~al.(2019{\natexlab{a}})Rostami, Kolouri, Eaton, and
  Kim}]{rostami2019deep}
Mohammad Rostami, Soheil Kolouri, Eric Eaton, and Kyungnam Kim.
  2019{\natexlab{a}}.
\newblock Deep transfer learning for few-shot sar image classification.
\newblock \emph{Remote Sensing}, 11(11):1374.

\bibitem[{Rostami et~al.(2020)Rostami, Kolouri, Pilly, and
  McClelland}]{rostami2020generative}
Mohammad Rostami, Soheil Kolouri, Praveen Pilly, and James McClelland. 2020.
\newblock Generative continual concept learning.
\newblock In \emph{Proceedings of the AAAI Conference on Artificial
  Intelligence}, volume~34, pages 5545--5552.

\bibitem[{Rostami et~al.(2019{\natexlab{b}})Rostami, Kolouri, and
  Pilly}]{rostami2019complementary}
Mohammad Rostami, Soheil Kolouri, and Praveen~K Pilly. 2019{\natexlab{b}}.
\newblock Complementary learning for overcoming catastrophic forgetting using
  experience replay.
\newblock In \emph{Proceedings of the 28th International Joint Conference on
  Artificial Intelligence}, pages 3339--3345. AAAI Press.

\bibitem[{Roweis(1998)}]{roweis1998algorithms}
Sam~T Roweis. 1998.
\newblock Em algorithms for pca and spca.
\newblock In \emph{Proceedings of NeurIPS}, pages 626--632.

\bibitem[{Ruder and Plank(2018)}]{ruder2018strong}
Sebastian Ruder and Barbara Plank. 2018.
\newblock Strong baselines for neural semi-supervised learning under domain
  shift.
\newblock In \emph{Proceedings of ACL)}, pages 1044--1054.

\bibitem[{Saito et~al.(2018)Saito, Ushiku, and Harada}]{saito2017ASYMmetric}
K.~Saito, Y.~Ushiku, and T.~Harada. 2018.
\newblock Asymmetric tri-training for unsupervised domain adaptation.
\newblock In \emph{ICML}.

\bibitem[{Sarma et~al.(2019)Sarma, Liang, and Sethares}]{sarma2019shallow}
Prathusha~Kameswara Sarma, Yingyu Liang, and William Sethares. 2019.
\newblock Shallow domain adaptive embeddings for sentiment analysis.
\newblock In \emph{Proceedings of EMNLP-IJCNLP}, pages 5552--5561.

\bibitem[{Shalev-Shwartz and Ben-David(2014)}]{shalev2014understanding}
Shai Shalev-Shwartz and Shai Ben-David. 2014.
\newblock \emph{Understanding machine learning: From theory to algorithms}.
\newblock Cambridge university press.

\bibitem[{Shen et~al.(2018)Shen, Qu, Zhang, and Yu}]{shen2018wasserstein}
Jian Shen, Yanru Qu, Weinan Zhang, and Yong Yu. 2018.
\newblock Wasserstein distance guided representation learning for domain
  adaptation.
\newblock In \emph{AAAI}.

\bibitem[{Stan and Rostami(2021{\natexlab{a}})}]{stan2021privacy}
Serban Stan and Mohammad Rostami. 2021{\natexlab{a}}.
\newblock Privacy preserving domain adaptation for semantic segmentation of
  medical images.
\newblock \emph{arXiv preprint arXiv:2101.00522}.

\bibitem[{Stan and Rostami(2021{\natexlab{b}})}]{stan2021unsupervised}
Serban Stan and Mohammad Rostami. 2021{\natexlab{b}}.
\newblock Unsupervised model adaptation for continual semantic segmentation.
\newblock In \emph{Proceedings of the AAAI Conference on Artificial
  Intelligence}, volume~35, pages 2593--2601.

\bibitem[{Sun and Saenko(2016)}]{sun2016deep}
Baochen Sun and Kate Saenko. 2016.
\newblock Deep coral: Correlation alignment for deep domain adaptation.
\newblock In \emph{European conference on computer vision}, pages 443--450.
  Springer.

\bibitem[{Tommasi and Caputo(2013)}]{tommasi2013frustratingly}
Tatiana Tommasi and Barbara Caputo. 2013.
\newblock Frustratingly easy nbnn domain adaptation.
\newblock In \emph{Proceedings of CVPR}, pages 897--904.

\bibitem[{Wang and Deng(2018)}]{wang2018deep}
Mei Wang and Weihong Deng. 2018.
\newblock Deep visual domain adaptation: A survey.
\newblock \emph{Neurocomputing}, 312:135--153.

\bibitem[{Wei et~al.(2018)Wei, Zhang, Gao, and Tian}]{wei2018person}
Longhui Wei, Shiliang Zhang, Wen Gao, and Qi~Tian. 2018.
\newblock Person transfer gan to bridge domain gap for person
  re-identification.
\newblock In \emph{Proceedings of the IEEE Conference on Computer Vision and
  Pattern Recognition}, pages 79--88.

\bibitem[{Wu and Huang(2016)}]{wu2016sentiment}
Fangzhao Wu and Yongfeng Huang. 2016.
\newblock Sentiment domain adaptation with multiple sources.
\newblock In \emph{Proceedings of ACL}, pages 301--310.

\bibitem[{Xi et~al.(2020)Xi, Zhuang, Zhou, Cheng, Lin, and He}]{xi2020domain}
Dongbo Xi, Fuzhen Zhuang, Ganbin Zhou, Xiaohu Cheng, Fen Lin, and Qing He.
  2020.
\newblock Domain adaptation with category attention network for deep sentiment
  analysis.
\newblock In \emph{Proceedings of The Web Conference 2020}, pages 3133--3139.

\bibitem[{Zellinger et~al.(2017)Zellinger, Grubinger, Lughofer,
  Natschl{\"a}ger, and Saminger-Platz}]{zellinger2017central}
Werner Zellinger, Thomas Grubinger, Edwin Lughofer, Thomas Natschl{\"a}ger, and
  Susanne Saminger-Platz. 2017.
\newblock Central moment discrepancy (cmd) for domain-invariant representation
  learning.
\newblock \emph{arXiv preprint arXiv:1702.08811}.

\bibitem[{Ziser and Reichart(2018)}]{ziser2018pivot}
Yftah Ziser and Roi Reichart. 2018.
\newblock Pivot based language modeling for improved neural domain adaptation.
\newblock In \emph{Proceedings of NAACL: Human Language Technologies, Volume
  1}, pages 1241--1251.

\bibitem[{Ziser and Reichart(2019)}]{ziser2019task}
Yftah Ziser and Roi Reichart. 2019.
\newblock Task refinement learning for improved accuracy and stability of
  unsupervised domain adaptation.
\newblock In \emph{Proceedings of ACL}, pages 5895--5906.

\end{thebibliography}
\bibliographystyle{acl_natbib}

\clearpage
\appendix

\section{Proof of Theorem~1}

We use the following theorem by Redko et al.~\cite{redko2017theoretical} and   a result by Bolley~\cite{bolley2007quantitative} on convergence of the empirical distribution to the true distribution in terms of the WD distance in our proof.

\textbf{Theorem~2 (Redko et al.~\cite{redko2017theoretical})}: Under the assumptions described in our framework, assume that a model is trained on the source domain, then for any $d'>d$ and $\zeta<\sqrt{2}$, there exists a constant number $N_0$ depending on $d'$ such that for any  $\bm{x}_i>0$ and $\min(N,M)\ge \max (\bm{x}i^{-(d'+2),1})$ with probability at least $1-\bm{x}_i$, the following holds:
\begin{equation}
\begin{split}
e_{\mathcal{T}}\le & e_{\mathcal{S}} +W(\hat{\mu}_{\mathcal{T}},\hat{\mu}_{\mathcal{S}})+e_{\mathcal{S},\mathcal{T}}\\& +  \sqrt{\big(2\log(\frac{1}{\bm{x}_i})/\zeta\big)}\big(\sqrt{\frac{1}{N}}+\sqrt{\frac{1}{M}}\big).
\end{split}
\label{eq:theroemfromcourty}
\end{equation}    
 
Theorem~2 provides an upperbound for the performance of a source-trained model in the target domain Redko et al.~\cite{redko2017theoretical} prove  Theorem~2 for a binary classification setting. We also provide  our proof in this case but it can be   extended.

The second term in Eq.~\eqref{eq:theroemfromcourty} demonstrates the effect of domain shift on the performance of a source-trained model in a target domain. When the distance between the two distributions is significant, this term will be large and hence the upperbound in Eq.~\eqref{eq:theroemfromcourty} will be loose which means potential performance degradation. Our algorithm mitigates domain gap because this term is minimized by minimization of  the second and the third terms  in Theorem~1.

\textbf{Theorem~1 }:   Consider that we the procedure described in Algorithm~1 for cross-domain sentiment analysis, then the following inequality  holds for the target expected error:
\begin{equation}
\small
\begin{split}
e_{\mathcal{T}}\le & e_{\mathcal{S}} +\hat{D}(\hat{\mu}_{\mathcal{S}},\hat{\mu}_{\mathcal{P}})+\hat{D}(\hat{\mu}_{\mathcal{T}},\hat{\mu}_{\mathcal{P}})+(1-\tau)\\&+e_{\mathcal{S},\mathcal{P}}+\sqrt{\big(2\log(\frac{1}{\bm{x}_i})/\zeta\big)}\big(\sqrt{\frac{1}{N}}+\sqrt{\frac{1}{M }}+2\sqrt{\frac{1}{N_p }}\big),
\end{split}
\label{eq:theroemforPLnips1}
\end{equation}    
where    $\bm{x}_i$ is a constant which depends on $\mathcal{L}(\cdot)$ and  $e_C'(\bm{w}^*)$  denotes the expected risk of the optimally joint trained model when used   on both the source domain and  the    pseudo-dataset.

\textbf{Proof:}   Due to the construction of the pseudo-dataset, the probability that the  predicted  labels for the pseudo-data points to be false  is equal to $1-\tau$.  Let:
\begin{equation}
\begin{split}
 & |\mathcal{L}(h_{\bm{w}_0}(\bm{z}^p_i),\bm{y}^p_i)- \mathcal{L}(h_{\bm{w}_0}(\bm{z}^p_i),\hat{\bm{y}}_i^{p})|=\\& \begin{cases}
    0, & \text{if $\bm{y}^t_i=\hat{\bm{y}}_i^{t}$}.\\
    1, & \text{otherwise}.
  \end{cases}
\end{split}
\label{eq:theroemforPLproof}
\end{equation}    
We use Jensen's inequality and take  expectation on both sides of \eqref{eq:theroemforPLproof} to deduce:
\begin{equation}
\begin{split}
|e_{\mathcal{P}}-e_{\mathcal{T}}|\le&\mathbb{E}\big(|\mathcal{L}(h_{\bm{w}_0}(\bm{z}^p_i),\bm{y}^p_i)- \mathcal{L}(h_{\bm{w}_0}(\bm{z}^p_i),\hat{\bm{y}}_i^{p})|\big)\\&\le 
(1-\tau).
\end{split}
\label{eq:theroemforPLproofexpectation}
\end{equation}    
Applying ~\eqref{eq:theroemforPLproofexpectation} in the below,   deduce:
\begin{equation}
\begin{split}
&e_{\mathcal{S}}+e_{\mathcal{T}}=e_{\mathcal{S}}+e_{\mathcal{T}}+e_{\mathcal{P}}-e_{\mathcal{P}}\le  \\&
e_{\mathcal{S}}+e_{\mathcal{P}}+|e_{\mathcal{T}}-e_{\mathcal{P}}|\le  
e_{\mathcal{S}}+e_{\mathcal{P}}+(1-\tau).
\end{split}
\label{eq:theroemforPLprooftrangleinq}
\end{equation}    
Taking infimum on both sides of ~\eqref{eq:theroemforPLprooftrangleinq}, we deduce:
\begin{equation}
\begin{split}
e_{\mathcal{S},\mathcal{T}}\le e_{\mathcal{S},\mathcal{P}}+(1-\tau).
\end{split}
\label{eq:theroemforPLprooftartplerror}
\end{equation}    
Now by considering
Theorem~2 for the two domains $\mathcal{S}$ and $\mathcal{T}$ and then using ~\eqref{eq:theroemforPLprooftartplerror}  in \eqref{eq:theroemfromcourty},  we can conclude:
\begin{equation}
\begin{split}
e_{\mathcal{T}}\le & e_{\mathcal{S}} +D(\hat{\mu}_{\mathcal{T}},\hat{\mu}_{\mathcal{S}})+e_{\mathcal{S},\mathcal{P}}+(1-\tau)\\& + \sqrt{\big(2\log(\frac{1}{\bm{x}_i})/\zeta\big)}\big(\sqrt{\frac{1}{N}}+\sqrt{\frac{1}{M}}\big).
\end{split}
\label{eq:theroemfromcourty1}
\end{equation}    
Now using the triangular inequality on the   metrics we can deduce:
\begin{equation}
\begin{split}
& D(\hat{\mu}_{\mathcal{T}},\hat{\mu}_{\mathcal{S}})\le  D(\hat{\mu}_{\mathcal{T}},\mu_{\mathcal{P}})+D(\hat{\mu}_{\mathcal{S}},\mu_{\mathcal{P}})\\&  \le D(\hat{\mu}_{\mathcal{T}},\hat{\mu}_{\mathcal{P}})+D(\hat{\mu}_{\mathcal{S}},\hat{\mu}_{\mathcal{P}})+2D(\hat{\mu}_{\mathcal{P}},\mu_{\mathcal{P}}).
\end{split}
\label{eq:theroemfromcourty2}
\end{equation}

  Now we   replace the term $D(\hat{\mu}_{\mathcal{P}},\mu_{\mathcal{P}})$ with its empirical counterpart using Theorem 1.1 in the work by  ~\cite{bolley2007quantitative}.

  \textbf{Theorem~3} (Theorem 1.1 by Bolley et al.~\cite{bolley2007quantitative}): consider that $p(\cdot) \in\mathcal{P}(\mathcal{Z})$ and $\int_{\mathcal{Z}} \exp{(\alpha \|\bm{x}\|^2_2)}dp(\bm{x})<\infty$ for some $\alpha>0$. Let $\hat{p}(\bm{x})=\frac{1}{N}\sum_i\delta(\bm{x}_i)$ denote the empirical distribution that is built from the samples $\{\bm{x}_i\}_{i=1}^N$ that are drawn i.i.d from $\bm{x}_i\sim p(\bm{x})$. Then for any $d'>d$ and $\bm{x}_i<\sqrt{2}$, there exists $N_0$ such that for any $\epsilon>0$ and $N\ge N_o\max(1,\epsilon^{-(d'+2)})$, we have:
 \begin{equation}
\begin{split}
P(W(p,\hat{p})>\epsilon)\le \exp(-\frac{-\bm{x}_i}{2}N\epsilon^2),
\end{split}
\label{eq:mainSuppICML3}
\end{equation}  
where $W$ denotes the WD distance. This relation measures the distance between the   empirical distribution and the true distribution, expressed in the WD distance.

 Applying  ~\eqref{eq:theroemfromcourty2} and ~\eqref{eq:mainSuppICML3} on ~\eqref{eq:theroemfromcourty1}  concludes Theorem~2 as stated:
\begin{equation}
\small
\begin{split}
e_{\mathcal{T}}\le & e_{\mathcal{S}} +D(\hat{\mu}_{\mathcal{S}},\hat{\mu}_{\mathcal{P}})+D(\hat{\mu}_{\mathcal{T}},\hat{\mu}_{\mathcal{P}})+(1-\tau)+e_{\mathcal{S},\mathcal{P}}+\\&\sqrt{\big(2\log(\frac{1}{\bm{x}_i})/\zeta\big)}\big(\sqrt{\frac{1}{N}}+\sqrt{\frac{1}{M }}+2\sqrt{\frac{1}{N_p }}\big),
\end{split}
\label{eq:theroemforPLnips55}
\end{equation}    

 \section{Complexity analysis for GMM estimation}
Estimating a GMM distribution usually is a computationally expensive tasks. The major reason is that normally the data points are unlabeled. This would necessitate relying on iterative algorithms such expectation maximization (EM) algorithm~\cite{moon1996expectation}. Preforming iterative E and M steps until convergence leads to high computational complexity~\cite{roweis1998algorithms}. However, estimating the multimodal distribution with  a GMM distribution is much simpler in our learning setting. Existence of labels helps us to decouple the Gaussian components    and compute the parameters using MAP estimate for each of the mode parameters in one step as follows:
\begin{equation}
\begin{split}
&\hat{\alpha}_j = \frac{|\bm{S}_j|}{N},\hspace{2mm}\\&\hat{\bm{\mu}}_j = \sum_{(\bm{x}_i^s,\bm{y}_i^s)\in \bm{S}_j}\frac{1}{|\bm{S}_j|}\phi_v(\bm{x}_i^s),\hspace{2mm} \\&\hat{\bm{\Sigma}}_j =\sum_{(\bm{x}_i^s,\bm{y}_i^s)\in \bm{S}_j}\frac{1}{|\bm{S}_j|}\big(\phi_v(\bm{x}_i^s)-\hat{\bm{\mu}}_j\big)^\top\big(\phi_v(\bm{x}_i^s)-\hat{\bm{\mu}}_j\big).
\end{split}
\label{eq:MAPestICLR1}
\end{equation}    
Given the above and considering that the source domain data is balanced, complexity of computing $\alpha_j$ is $O(N)$ (just checking whether data points $\bm{x}_i^s$ belong to  $\bm{S}_j$).  Complexity of computing $\bm{\mu}_j$ is $O(NF/k)$, where $F$ is the dimension of the embedding space. Complexity of computing the co-variance matrices $\bm{\Sigma}_j$ is $O(F(\frac{N}{k})^2)$. Since, we have $k$ components, the total complexity of computing GMM is $O(\frac{FN^2}{k})$. If $O(F)\approx O(k)$, which seems to be a reasonable practical assumption, then the total complexity of computing GMM would be $O(N^2)$. Given the large number of learnable parameters in most deep neural networks which are more than $N$ for most cases, this complexity is fully dominated by complexity of a single step of backpropagation. Hence, this computing the GMM parameters does not increase the computational complexity for.



\end{document}